\documentclass[sigconf]{acmart}
\AtBeginDocument{%
  \providecommand\BibTeX{{%
    \normalfont B\kern-0.5em{\scshape i\kern-0.25em b}\kern-0.8em\TeX}}}

\setcopyright{acmcopyright}
\copyrightyear{2018}
\acmYear{2018}
\acmDOI{XXXXXXX.XXXXXXX}

\acmConference[Conference acronym 'XX]{Make sure to enter the correct
  conference title from your rights confirmation emai}{June 03--05,
  2018}{Woodstock, NY}
\acmPrice{15.00}
\acmISBN{978-1-4503-XXXX-X/18/06}

\begin{document}

\title{Ethereum Fraud Detection with \\ Heterogeneous Graph Neural Networks}

\author{Hiroki Kanezashi}
\authornote{Both authors contributed equally to this research.}
\authornote{A part of this work was done while this author was in AIST.}
\email{hkanezashi@ds.itc.u-tokyo.ac.jp}
\author{Toyotaro Suzumura}
\authornotemark[1]
\email{suzumura@acm.org}
\affiliation{%
  \institution{The University of Tokyo}
  \city{Tokyo}
  \country{Japan}
}

\author{Xin Liu}
\author{Takahiro Hirofuchi}
\affiliation{%
  \institution{National Institute of Advanced Industrial Science and Technology (AIST)}
  \city{Tokyo}
  \country{Japan}}

\begin{abstract}
While transactions with cryptocurrencies such as Ethereum are becoming more prevalent, fraud and other criminal transactions are not uncommon. Graph analysis algorithms and machine learning techniques detect suspicious transactions that lead to phishing in large transaction networks. Many graph neural network (GNN) models have been proposed to apply deep learning techniques to graph structures. Although there is research on phishing detection using GNN models in the Ethereum transaction network, models that address the scale of the number of vertices and edges and the imbalance of labels have not yet been studied. In this paper, we compared the model performance of GNN models on the actual Ethereum transaction network dataset and phishing reported label data to exhaustively compare and verify which GNN models and hyperparameters produce the best accuracy. Specifically, we evaluated the model performance of representative homogeneous GNN models which consider single-type nodes and edges and heterogeneous GNN models which support different types of nodes and edges. We showed that heterogeneous models had better model performance than homogeneous models. In particular, the RGCN model achieved the best performance in the overall metrics.
\end{abstract}

\ccsdesc[100]{Social and professional topics~Financial crime}
\ccsdesc[500]{Mathematics of computing~Graph theory}
\ccsdesc[500]{Computing methodologies~Neural networks}
\ccsdesc[500]{Computing methodologies~Supervised learning by classification}

\keywords{graph neural networks, graph convolutional networks, heterogeneous graph, cryptocurrency, phishing detection}

\maketitle
\section{Introduction}

\subsection{Background and Problem Statement}
Blockchain is a distributed public database that allows data exchange between two parties without centralized authority. Recently, various applications using Blockchain technologies have been proposed and are already in practice on large platforms such as Bitcoin and Ethereum. However, money laundering, phishing scams, and other financial crimes are common within these platforms and difficult to uncover due to the many transactions, which will be discussed later.

Each transaction through traditional banks is constantly monitored and strictly checked when a new account is opened and large suspicious transactions are held. However, it is easy to open accounts (create "wallets") and make transactions freely without the need for a third party, making it difficult to check for suspicious transactions in advance. There is concern that transactions that lead to crime could be hidden. On the other hand, all transaction details are public due to the nature of the blockchain, and anyone can obtain transaction information. However, as cryptocurrency transactions become more widespread, the number of transactions and wallets increase, making detection time-consuming.

Many studies have been conducted to detect suspicious transactions in such large financial transaction networks. Most of these studies have modeled the patterns of fraudulent transactions based on the timestamps and amounts of one or more transactions as features and detected unknown transactions with these models. Many studies have conducted model optimization and evaluation of fraud detection using graph embedding and incidental feature fees, but few have been conducted using GNN models.

Recently, a graph convolution algorithm based on deep learning techniques has been proposed to automatically generate such features using a graph algorithm rather than manually.
The Graph Convolutional Model (GCN)\cite{gcn} model and its derivative models that apply neural network models to graph data have been proposed in recent years. They learn by updating the feature values of each node and edge according to the graph structure. In addition, some models incorporate the information associated with the vertices and edges of the graph directly into the learning process. For example, the Graph Attention Network (GAT)\cite{gat} learns the importance of vertices in the form of attention. The Relational GCN (RGCN)\cite{rgcn} learns parameters according to the role of the edges (edge type); thus, learning different types of it is possible to perform more informative learning on heterogeneous graphs. Despite the wide variety of proposed GNN models, few have been applied to detect fraudulent transactions on cryptocurrency trading networks. It is still unclear which GNN models are effective in practice.

\subsection{Our Contributions}

This study will thoroughly examine which GNN models and hyperparameters effectively detect phishing fraud transactions from public Ethereum transaction networks. First, we construct a heterogeneous transaction network that incorporates the basic homogeneous GNN models such as GCN\cite{gcn}, GraphSAGE\cite{graphsage}, and GAT\cite{gat} models and node types such as accounts (wallets and exchanges) in the transaction network data. Then we developed a model performance compared to heterogeneous GNN models such as RGCN\cite{rgcn} and HGT\cite{hgt}, which correspond to heterogeneous graphs.

The main contributions of our work are as follows.
\begin{enumerate}
    \item We investigate the characteristics of a real Ethereum transaction network, and we found some types of nodes play the role of hubs in the network.
    \item Based on the statistical evaluations, we evaluated and compared the model performance of GNN models with the transaction network as a heterogeneous graph.
    \begin{enumerate}
        \item We conduct exhaustive evaluations of model performances with various graph neural network (GNN) models based on these hypotheses.
        \item We compare the model performance of GNN models and hyperparameters.
        \item Mainly, we focus on the heterogeneity of its network, which each node and edge in the Ethereum transaction network has a type.
    \end{enumerate}
\end{enumerate}

The remainder of this paper is as follows. In Section 2, we present related work on the problem of detecting fraudulent transactions in finance and methods using graph convolution and GNN models. In Section 3, we introduce representative homogeneous and heterogeneous GNN models, describe the characteristics of each model, and present hypotheses for phishing fraud detection applications. In Section 4, we construct a transaction network from actual Ethereum transaction logs and phishing fraud report data, evaluate the model performance of each GNN model, and compare the results. In Section 5, we test and discuss how correct our hypotheses in Section 3 are in light of the results in Section 4. Finally, Section 6 concludes this study and discusses future work.

\section{Related Work}

Fraud detection in the Ethereum transaction network is one of the hot topics due to its social importance and the availability of public datasets. Financial transaction networks usually have edge attributes: timestamp and amount. These attributes are assumed to be the key to fraud detection. Many fraud detection methods have been proposed using transaction edge attributes based on this assumption. However, some proposed models and algorithms are used with arbitrary features, which may not apply to other transaction network datasets or other types of fraud detection tasks, such as money laundering detection.

With the development of graph embedding and graph neural network research, many phishing detection methods using graph embeddings have been proposed. The advantage of using graph embedding and GNN models is to capture features of suspicious account node features.

Wu et al. proposed a transaction embedding model named trans2vec\cite{trans2vec}, incorporating the amount and timestamp properties of the transaction edges. They compared trans2vec with state-of-the-art embedding algorithms, and it achieved better model performance with time and amount of features. While creating account node-based embedding vectors, it applies random walks with the amount and time-biased sampling.

Chen et al. proposed another phishing detection model\cite{acmtran} based on GCN and autoencoder. It samples subgraphs by random walk and applies node embeddings and a GCN model to incorporate spatial structures and node features. The proposed model performed better than other traditional embedding methods. However, some node features from transaction data are determined arbitrarily, and only the GCN model is used to extract the structural features.

Lin et al. modeled the Ethereum transaction network as a weighted temporal graph and a Temporal Weighted Multidigraph Embedding (T-EDGE)\cite{t-edge} to incorporate temporal and weighted transaction edges. In T-EDGE, it extends the edge probabilities to be visited in the random walk by transaction amount and interval.
Xie et al. also modeled it as a temporal-amount snapshot multigraph (TASMG) and extended the random walk named temporal-amount walk (TAW)\cite{tasmg} to generate embeddings.
Construct a transaction subgraph network (TSGN)\cite{tsgn} with a learnable transaction weight mapping function and directed-TSGN to be aware of the edges. The model performance of (Directed-)TSGN is better than the baseline method in other Ethereum datasets.

\section{Graph Neural Networks}
Graph neural networks (GNN) are neural network models that capture the structure of the graph by message passing between the nodes in a graph. Many derived GNN models have been proposed, including graph convolution, graph attention, and heterogeneous GNN, depending on the graph data used and the application. This section describes representative homogeneous GNN and heterogeneous GNN models and hypothesizes their effectiveness in phishing scam detection applications using Ethereum networks, respectively.

\subsection{Homogeneous GNN Models}
First, we use GNN models for homogeneous graphs: GCN (graph convolutional network), GAT (graph attention network), and GraphSAGE.

GCN (graph convolutional network)\cite{gcn} is a neural network model based on its graph structure data that continuously accumulates the node's neighbor's feature vectors, feeds it to a neural network, and applies nonlinear functionality. The GCN model does not consider transaction properties (count and amount) as "edge weights," which may prevent detecting suspicious accounts with many transactions. No matter how many transactions are made, all edge weights in the GCN model are the same.

In phishing detection applications in Ethereum networks, not all neighboring account nodes are related to suspicious transactions or fraudulent neighboring accounts. Each node has the same weights as all neighboring nodes in the GCN model. On the other hand, the GAT (graph attention network) model \cite{gat} implicitly learns the different weight of each neighborhood as "attention". We may distinguish normal and suspicious accounts more effectively with the attention mechanism.

The original GCN model cannot be applied to large graph structures because of its computational cost to learn embeddings for all nodes. The GraphSAGE\cite{graphsage} model learns the aggregation function from neighboring nodes instead of embedding nodes. The most vital points of GraphSAGE are that the learned aggregation function is applicable for unknown nodes and reduces the computational functions for large graphs with neighboring node samplings.

\subsection{Heterogeneous GNN Models}
In Ethereum transaction networks, some nodes have different attributes such as "exchanges" and "wallets" as well as normal accounts. In addition, vertices representing accounts. Such graphs with different vertices and edges are called heterogeneous graphs, which is more informative for machine learning and other graph analysis than a homogeneous graph with the usual types. In machine learning using GNN models, GNN models corresponding to many heterogeneous graphs (heterogeneous GNN) have been proposed because of their increasing application to knowledge graphs and other graphs.

Although our research aims to detect accounts that commit phishing fraud, it can be expected that including nodes with other roles, such as exchanges, will detect suspicious accounts with greater accuracy. This section describes the representative heterogeneous GNN models used in the evaluation experiments: the RGCN, HAN, and HGT models, respectively.

Kipf et al. proposed the Relational GCN (RGCN)\cite{rgcn} model, a generalization GCN model that accepts multi-relational graphs equivalent to edge types. In the RGCN layers, the weight parameters of each relation and self-loop are separately trained, while the original GCN model trains all of connecting edges uniformly.

As the heterogeneity extension of the GAT model, Wang et al. proposed a Heterogeneous Graph Attention network (HAN)\cite{han}. This model learns two types of attention (node-level and semantic-level attention) from the sequence of different node types called "meta-path". The node-level attention learns importance between nodes in each meta-path, and the semantic-level attention learns the importance of different meta-paths.

Inspired by the architecture design of the Transformer\cite{transformer} attention mechanism, Hu et al. proposed the Heterogeneous Graph Transformer (HGT)\cite{hgt} model, which models large-scale dynamic heterogeneous graphs. Like the HAN model, HGT uses meta-relations (triplets of source, edge, and target types) to distinguish different type relationships. Even though temporal information can be incorporated by relative temporal encoding (RTE) to handle the temporal nature of web-scale graphs, RTE has not yet been implemented in this time.

\section{Experiments}

\subsection{Overview of GNN Tasks}
There are several possible approaches to financial fraud detection tasks, such as phishing fraud detection, including edge classification, which determines whether the transaction itself is fraudulent, node classification, which determines whether the user or account that made the transaction is fraudulent. Furthermore, subgraph classification determines whether the set of accounts and transactions is fraudulent. We employed the node classification method to determine fraudulent accounts in this work. In the original Ethereum trading network dataset, information on phishing fraud accounts was attached to the accounts (addresses in Ethereum). It is intuitive to use node classification for each account, exchange, and other corresponding vertices, as described below.

To evaluate the performance of each GNN model model and its hyperparameters, we adopt precision, recall, F1-score, and PR-AUC (area under the precision-recall curve) as performance metrics. As is typical in fraud detection with imbalanced labels, recall and PR-AUC are more critical because the main objective is to detect phishing accounts. Because of imbalanced datasets (very few phishing accounts), we adopt PR-AUC and precision, recall, and F1-score.

\subsection{Hardware and Software Setup}
In our experiments, we used GPUs and PyTorch as suitable hardware and libraries for deep learning to train exhaustively various GNN models and compare and evaluate their performance. To implement the various GNN models in a unified manner as generic neural network models, we also used the Deep graph library (DGL)\cite{dgl} as the GNN framework.

As the computation environment, we used NVIDIA DGX-1\cite{dgx1}. Although it has 8 NVIDIA Tesla GPU processors, we assigned one GPU to each GNN training process. Table \ref{tb:hardware} is a summary of DGX-1.

\begin{table}[ht]
\caption{Overview of DGX-1}
\label{tb:hardware}
\centering
\begin{tabular}{l|l}
\hline
GPU & 8X NVIDIA\textregistered Tesla\textregistered V100, 16GB \\
CPU & Intel\textregistered Xeon\textregistered CPU E5-2698 v4 2.20GHz, 2 x 20 cores \\
DRAM & 512 GB 2,133 MHz DDR4 RDIMM \\
\hline
\end{tabular}
\end{table}

Table \ref{tb:software} is the list of libraries for our experiments. We used CUDA version 11 and PyTorch 1.7 as libraries to train the GNN models on the GPUs. Both of the GNN frameworks DGL and PyG work on the GPU-bound PyTorch.

\begin{table}[ht]
\caption{Libraries and Softwares}
\label{tb:software}
\centering
\begin{tabular}{l|l}
\hline
Host OS & Ubuntu 18.04.4 LTS \\
Docker & 19.03.15, build 99e3ed8919 \\
GPU Driver & 418.197.02 \\
CUDA & 11.0 \\
GCC & 7.5.0 \\
Python & 3.6.9 \\
PyTorch & 1.7.1+cu110 \\
Deep Graph Library (DGL) & 0.7.1 \\
\hline
\end{tabular}
\end{table}

\subsection{Ethereum Transaction Network Datasets}

\subsubsection{Data Source and Subgraph Constructions}
As the real Ethereum transaction network dataset, we obtained a public transaction graph dataset from Kaggle\footnote{https://www.kaggle.com/xblock/ethereum-phishing-transaction-network}.
It contains 2,973,489 nodes, 13,551,303 edges, and 1,165 labeled (fraud) nodes. Each edge represents a single edge and has two attributes, the amount (ETH) and the timestamp.

Due to the imbalance of the node label, we extract a subgraph from the transaction network using a similar approach to \cite{trans2vec}. First, we picked up 2,230 nodes from the network (all 1,165 fraud nodes and 1,165 normal nodes chosen randomly) and extract incoming and outgoing neighboring nodes from all the chosen nodes. Then, we extract a subgraph that only consists of these chosen nodes and their neighbors. After subgraph extraction, the nodes and edges are 9,629 and 386,612, respectively.

\subsubsection{Extension to Heterogeneous Graphs}

Some nodes in this transaction network dataset have different roles from the ordinal account (address). While most nodes are ordinal accounts, a few nodes represent different types such as exchange accounts, token contracts, etc. Although these account node types are not directly related to phishing fraud detection applications because they cannot be labeled "fraud accounts," we assume that the information will be hints of suspicious surrounding transactions and account detection.

We add heterogeneity to the transaction graph and apply heterogeneous GNN models to the heterogeneous transaction graph based on this assumption. Since the transaction network itself does not contain node type, we obtained the role of nodes from another data source named Etherscan\footnote{https://etherscan.io/labelcloud}.
Although there are hundreds of labels registered for account nodes on this website, we used the following labels, which may have a relationship with phishing transactions: exchange, token-contract, exchange, gaming, gambling, ico-wallets, wallet-app, cold-wallet. Then, we add labels as node attributes of our transaction graph. Note that not all nodes in our graph data have the corresponding node labels. We added a default label named "account" to other nodes. Table \ref{tb:nodetype} shows the number of vertices by node type.

\begin{table}[ht]
\caption{The number of nodes by type}
\label{tb:nodetype}
\centering
\begin{tabular}{l|r|r}
\hline
Node Type & Number of nodes & Percentage \\ \hline
account & 9167 & 95.20 \\
token-contract & 414 & 4.30  \\
exchange & 30 & 0.31\\
gaming-tokens & 9 & 0.09  \\
ico-wallets & 5 & 0.05 \\
wallet-app & 2 & 0.02 \\
cold-wallet & 1 & 0.01 \\
gambling-accounts & 1 & 0.01 \\ \hline
Total & 9629 & 100.00 \\
\hline
\end{tabular}
\end{table}

\begin{figure}[htb]
    \centering
    \includegraphics[width=0.8\hsize]{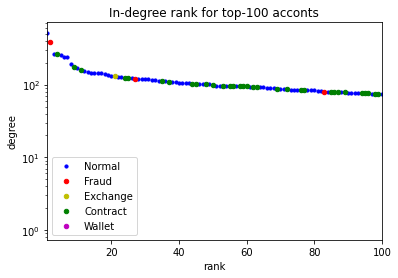}
    \caption{In-degree of the Top 100 nodes by type}
    \label{fig:in-hub}
\end{figure}

\begin{figure}[htb]
    \centering
    \includegraphics[width=0.8\hsize]{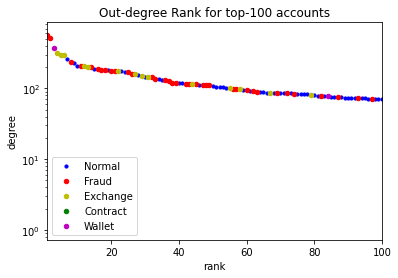}
    \caption{Out-degree of the Top 100 nodes by type}
    \label{fig:out-hub}
\end{figure}

Figure \ref{fig:in-hub} and Figure \ref{fig:out-hub} show the trends of in-degree and out-degree of the top 100 hub nodes, respectively. Most of the hub vertices with higher in-degrees are normal accounts, but some "contract" nodes also have many incoming edges. On the other hand, many "exchange" nodes and fraud (phishing) accounts have large out-degree. Even the "account" nodes consist of more than 95\% of all nodes, the "exchange" and "token-contract" nodes are connected to many other nodes, which will be the key to fraud account detection.

We added edge types based on the source and target edge types to adopt heterogeneous GNN models in the next step. Many heterogeneous GNN models consider edge types rather than node types. For example, The RGCN model applies graph convolution for each edge type separately, and the HAN model learns heterogeneous node relationships with meta-paths (sequences of node and edge types). We concatenated the types of the source and target nodes into a new label as the edge type. For example, an edge connecting from the "account" type node to "wallet-app" has the type "account - wallet-app." Finally, we created 35 types of heterogeneous edges based on the following combination of the source and target node types in Table \ref{tb:edgetype}.

\begin{table*}[htb]
    \caption{The number of Edges by Type}
    \label{tb:edgetype}
    \centering
    \begin{tabular}{|r|r|r|r|r|r|r|r|r|}
    \hline
        \multicolumn{1}{|c|}{Source} &  \multicolumn{8}{|c|}{Target} \\ \hline
        ~ & Account & exchange & token-contract & wallet-app & ico-wallets & cold-wallet & gaming & gambling \\ \hline
        Account & 8690075 & 265718 & 3274017 & 18588 & 64123 & 9952 & 83230 & 20730 \\ \hline
        exchange & 667308 & 39111 & 328130 & 8 & 1594 & 0 & 7961 & 0 \\ \hline
        token-contract & 35465 & 7 & 0 & 3 & 17143 & 0 & 0 & 0 \\ \hline
        wallet-app & 6556 & 7 & 189 & 0 & 14 & 0 & 0 & 0 \\ \hline
        ico-wallets & 8507 & 1 & 4389 & 0 & 2 & 0 & 0 & 0 \\ \hline
        cold-wallet & 23 & 0 & 7 & 62 & 0 & 0 & 0 & 0 \\ \hline
        gaming & 14 & 0 & 0 & 0 & 0 & 0 & 0 & 0 \\ \hline
        gambling & 8357 & 1 & 0 & 11 & 0 & 0 & 0 & 0 \\ \hline
    \end{tabular}
\end{table*}

\begin{figure}[htb]
    \centering
    \includegraphics[width=0.9\hsize]{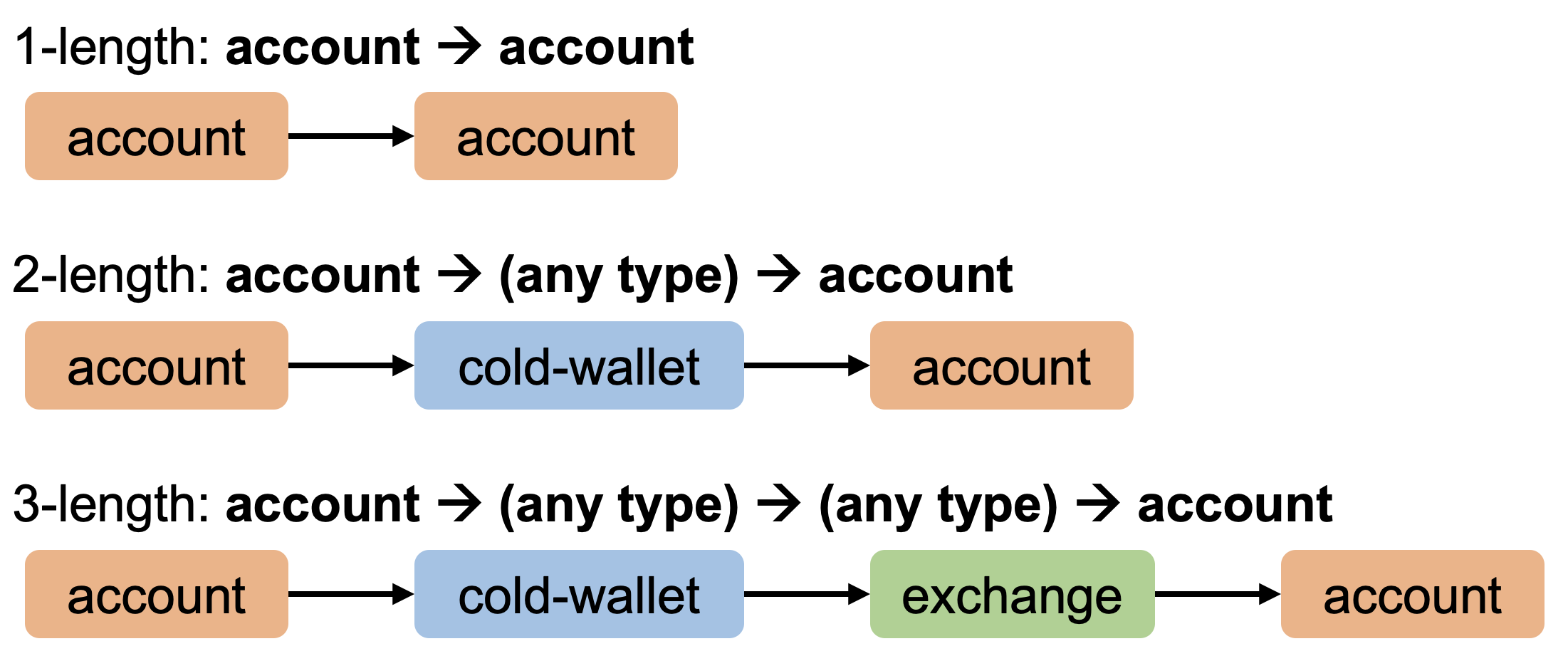}
    \caption{Examples of meta-paths in HAN}
    \label{fig:meta-path}
\end{figure}

In the same way as edge type labels, we also defined "meta-paths" for the HAN model. Our experiments apply the HAN model to meta-paths connected to "Account" nodes and perform classification only on the "Account" nodes with up to three edges. Because of the regulation of the HAN model, the meta-path types of the source and destination nodes must be the same as in Figure \ref{fig:meta-path}. As a result, the total number of meta-paths connected to "Account" node types is 49,103 (82.8\%).

Our target problem is to identify whether an account is fraudulent or not, so we are defining meta-paths where the source and target vertices should be "Accounts." Based on our statistical analysis on meta-paths, more meta-paths are related to "taken-contract," "exchange," or "wallet."

\begin{figure}[htb]
    \centering
    \includegraphics[width=\hsize]{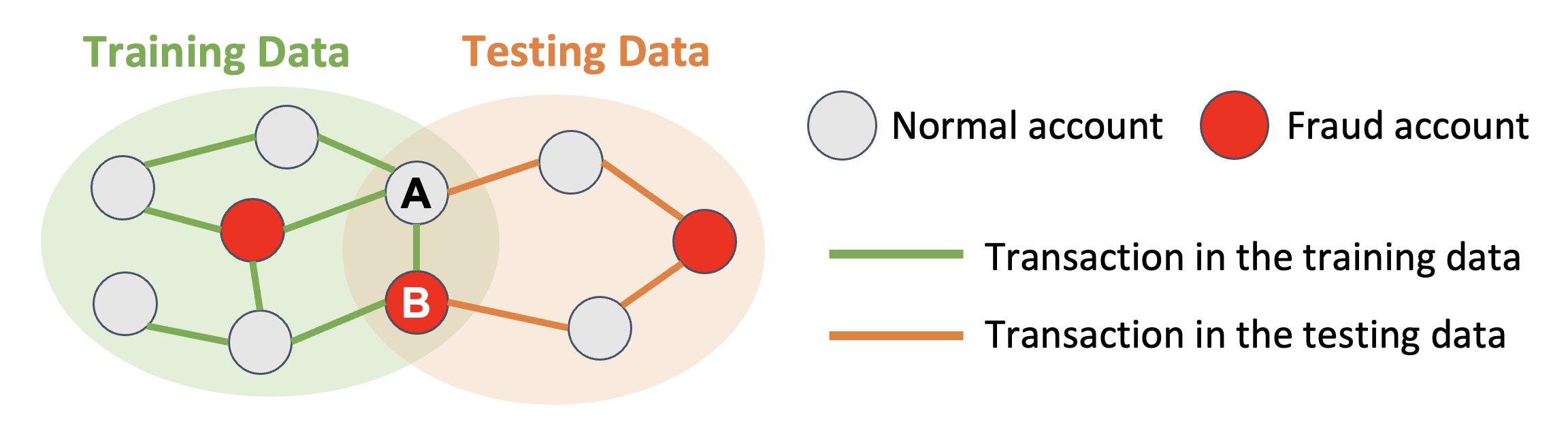}
    \caption{An Example of Nodes Involved in Training and Testing Datasets}
    \label{fig:data-split}
\end{figure}

For machine learning in financial transaction networks, the timestamps of the edges that represent transactions are meaningful. We split the transaction network into training and testing subgraphs by transaction timestamp. The training subgraph consists of 80\% former transactions, and the testing subgraph consists of 20\% latter transaction edges. We apply a node classification task for fraud account detection, so we must assign each node to training and testing data. However, by splitting the graph dataset with the transaction timestamps, some accounts (A and B in figure \ref{fig:data-split}) involved in both training and testing the transaction edges. In this case, fraud account B in both training and testing datasets turns out to be automatically determined as fraud (true positive) because it is already known as fraud. On the other hand, normal account A will be predicted as normal (true negative) or fraud (false positive) because it might become a fraud.

As input for each GNN model, we define the node feature set in Table \ref{tb:feats}. Some features are computed from the transaction amount and frequency for each node, and other features are from graph analytics (PageRank and degree distributions).

\begin{table}[ht]
\caption{Base Node Features}
\label{tb:feats}
\centering
\begin{tabular}{l|l}
\hline \hline
Feature Name & Description \\ \hline
send\_num & Transaction count sent from this account node \\
recv\_num & Transaction count received by this account node \\
send\_amount & Total amount sent from this account node \\
recv\_amount & Total amount received by this account node \\
in\_degree & Neighboring accounts for incoming transactions \\
out\_degree & Neighboring account for outgoing transactions \\
pagerank & PageRank score in this account node \\
\hline
\end{tabular}
\end{table}

\begin{figure}[htb]
    \centering
    \includegraphics[width=\hsize]{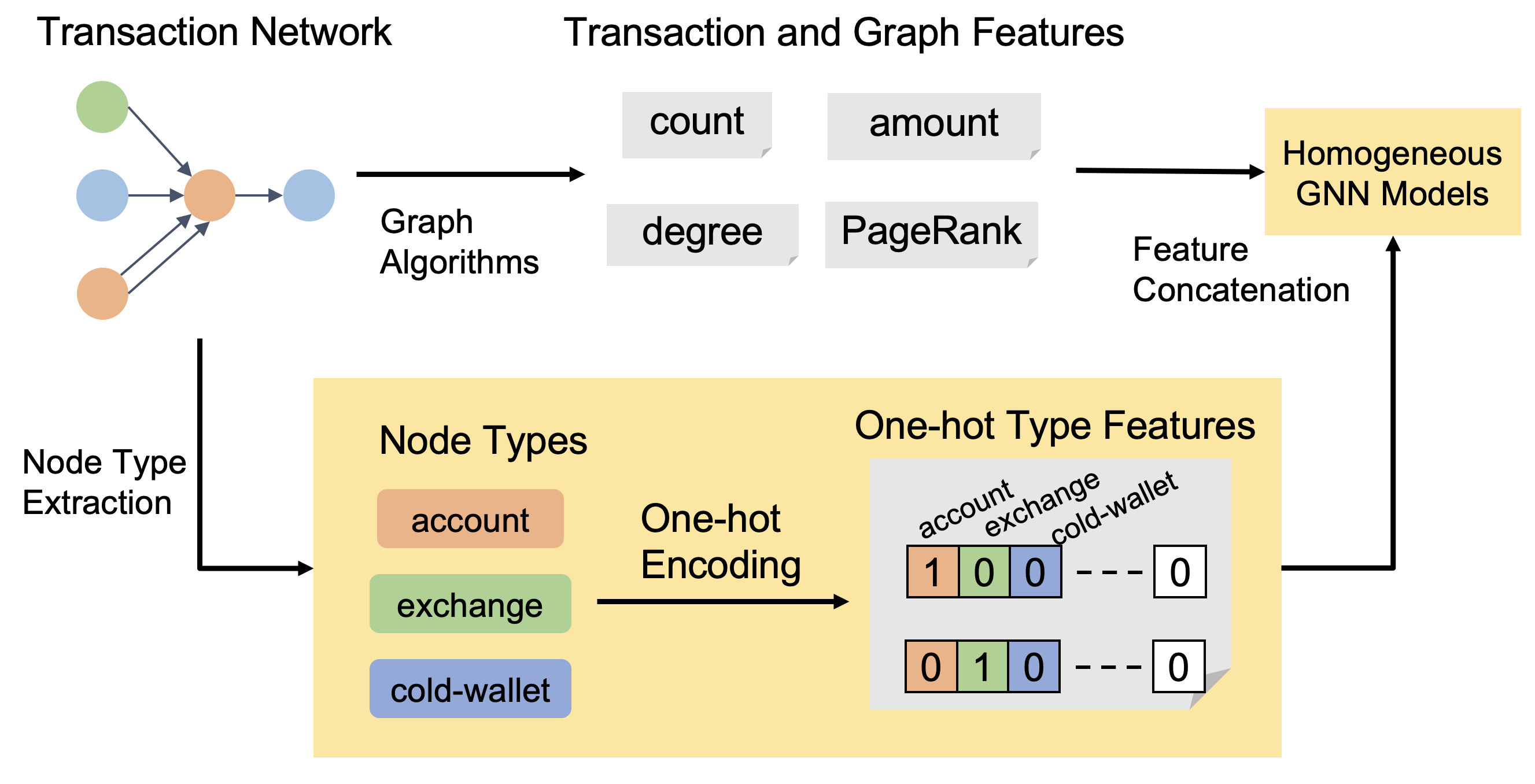}
    \caption{One-hot Node Type Features for Homogeneous GNN Models}
    \label{fig:one-hot}
\end{figure}

In addition to these account-based node features, we also define one-hot node type features to embed heterogeneous graph information to homogeneous GNN models. Figure \ref{fig:one-hot} describes how we convert the information from the node types into the input node features. We encode a one-hot vector for each node, which has only a nonzero value, and the others are zero corresponding node types. We also use these one-hot features in addition to the transaction and graph features in Table \ref{tb:feats} to evaluate the effectiveness of heterogeneous GNN models compared to a homogeneous model with node-type information.

\subsection{Model Performance Results}

\subsubsection{Homogeneous vs Heterogeneous GNN Models}

Table \ref{tb:model-perf} shows homogeneous (GCN, GAT, and GraphSAGE) model performance and heterogeneous (RGCN, HAN, HGT with/without temporal encoding) GNN models. The best score for each metrics is indicated in bold.

\begin{table}[htb]
    \caption{Model Performance of Homogeneous and Heterogeneous GNN Models}
    \label{tb:model-perf}
    \centering
    \begin{tabular}{l|r|r|r|r}
    \hline
        Model & Precision & Recall & F1 & PR-AUC \\ \hline
        GCN & 0.8817 & 0.4387 & 0.5843 & 0.7154 \\ 
        GAT & 0.8695 & 0.5093 & 0.6399 & 0.7558 \\ 
        SAGE & 0.8244 & 0.6644 & 0.7354 & 0.8150 \\ \hline
        HAN & 0.8800 & 0.4259 & 0.5725 & 0.6561 \\ 
        HGT & 0.7718 & \textbf{0.7940} & 0.7849 & 0.8383 \\ 
        RGCN & \textbf{0.8958} & 0.7454 & \textbf{0.8124} & \textbf{0.8896} \\ \hline
    \end{tabular}
\end{table}

Among homogeneous GNN models (GCN, GAT, and GraphSAGE), the GraphSAGE model achieves the best performance in recall, F1 score, and PR-AUC values.
On the other hand, although GCN achieved 88\% in precision, its recall was the lowest and results in only 58\% in F1-score. Among the heterogeneous GNN models (RGCN, HAN, and HGT), the RGCN model achieved the best precision, F1-score, and PR-AUC values. The HGT model achieved the best recall value (79\%) among the GNN models. However, in the HAN model, the overall model performance is almost the same as that of the baseline GCN model. Overall, we conclude that some heterogeneous models improve model performance and that the RGCN model produces the best model performance in these GNN models.

\subsubsection{Heterogeneous Graph and GNN Models}

With the result of the model performance among homogeneous and heterogeneous GNN models, we found that heterogeneous GNN models using heterogeneous edge types are better than homogeneous models. To confirm whether the better performance comes from the heterogeneous transaction network or heterogeneous GNN models, we added information of node types to the input node features as one-hot node type features.

Table \ref{tb:onehot-homogeneous} is the result of homogeneous GNN models with and without one-hot input features. The precision and PR-AUC have improved a little in the GCN and GAT models, but there is a slight improvement in the GraphSAGE model.

\begin{table}[htb]
    \caption{Model Performance of Homogeneous GNN Models with One-hot Features}
    \label{tb:onehot-homogeneous}
    \centering
    \begin{tabular}{l|r|r|r|r}
    \hline
        Model & Precision & Recall & F1 & PR-AUC \\ \hline
        GCN Base & 0.8819 & 0.4201 & 0.5711 & 0.7108 \\ 
        GCN One-hot & 0.9057 & 0.4560 & 0.6060 & 0.7265 \\ \hline
        GAT Base & 0.8630 & 0.5278 & 0.6562 & 0.7522 \\ 
        GAT One-hot & \textbf{0.9292} & 0.4838 & 0.6370 & 0.7702 \\ \hline
        SAGE Base & 0.8329 & \textbf{0.6620} & 0.7342 & 0.8147 \\ 
        Sage One-hot & 0.8284 & 0.6609 & \textbf{0.7370} & \textbf{0.8151} \\ \hline
    \end{tabular}
\end{table}

Figure \ref{tb:onehot-rgcn} compares homogeneous GNN models (GCN, GAT, and GraphSAGE) with one-hot node type features and the best heterogeneous GNN model (RGCN). Although the precision value of GCN and GAT models is better than the RGCN model, other metrics (recall, F1-score, and PR-AUC) of the RGCN model are still better than the other models.
Therefore, the RGCN model with a heterogeneous transaction network is the best for phishing account detection.

\begin{table}[htb]
    \caption{Model Performance of One-hot Features and RGCN Model}
    \label{tb:onehot-rgcn}
    \centering
    \begin{tabular}{l|r|r|r|r}
    \hline
        Model & Precision & Recall & F1 & PR-AUC \\ \hline
        GCN One-hot & 0.9057 & 0.4560 & 0.6060 & 0.7265 \\ 
        GAT One-hot & \textbf{0.9292} & 0.4838 & 0.6370 & 0.7702 \\ 
        SAGE One-hot & 0.8284 & 0.6609 & 0.7370 & 0.8151 \\ \hline
        RGCN & 0.8958 & \textbf{0.7454} & \textbf{0.8124} & \textbf{0.8896} \\ \hline
    \end{tabular}
\end{table}

\subsection{Hyperparameter Tuning}

In addition to the types of GNN models, we also examined how the model performance varies with different hyperparameters. The hyperparameters compared in this study are as follows: GAT heads, learning rate, hidden layer size, and dropout ratio.

\textbf{Attention Heads (GAT model only)}. The GAT model supports multi-head attention to introduce independent attention mechanisms. The more attention heads enable the GAT model to express more complex the relationships between nodes.

\textbf{Learning Rate}. Learning rate, a hyperparameter for weighting the error between the prediction results and the teacher data, is fed back to the GNN model by the error backpropagation method, as in learning other neural network models. The larger the learning rate, the shorter the time required to reach the optimal solution with fewer epochs, but if the learning rate is too large, the learned model becomes unstable and may be far from optimal.

\textbf{Hidden Layer Size}. In the GCN model, adding one more hidden layer to the GCN model results in the following convolution vector of the node itself can be propagated one more hop to the next node, allowing for learning that takes the graph structure into account. As with the hidden layers described above, the number of hidden units, or the number of training parameters per hidden layer, also changes the expressive power of the neural network.

\textbf{Dropout}. An essential technique to improve the accuracy of neural network models is dropout\cite{dropout}. Dropout suppresses overfitting by randomly disabling a certain percentage of hidden units during training. The higher the dropout rate, the smaller the proportion of hidden units to be trained, but there is a trade-off in that more training data and time are required to train the entire GNN.

\textbf{Weight Decay}. Another technique to reduce overfitting is weight decay, which adds the L2-norm to the backward process. Weight decay can suppress changes in drastic weight updates by adding L2 regularization when updating the weight parameters. The larger the value of this hyperparameter, the more the change is suppressed.

The candidates of hyperparameters are as Table\ref{tb:params}. The default values are indicated in bold.

\begin{table}[ht]
\caption{Candidates of Hyperparameters}
\label{tb:params}
\centering
\begin{tabular}{l|l}
\hline
Attention Heads & \textbf{2}, 3, 4 \\
Learning Rate & 0.001, 0.002, \textbf{0.005}, 0.010, 0.020, 0.050, 0.100 \\
Hidden Units & 16, 32, 64, \textbf{128}, 256 \\
Dropout Ratio & 0.00, 0.25, \textbf{0.50}, 0.75 \\
Weight Decay & 0.0000, 0.0001, \textbf{0.0005}, 0.0010, 0.0050 \\
\hline
\end{tabular}
\end{table}

We compared the model performances by the number of attention heads only for the GAT model. We used the GraphSAGE model for other hyperparameters, which achieves the best model performance among homogeneous GNN models.


Table \ref{tb:gat-heads} shows the model performance results of the GAT model with 2 (default), 3 and 4 attention heads and GCN and GraphSAGE models for comparison. With four attention heads, the recall and F1-score improved a little. However, the GraphSAGE model still keeps the best model performance in these overall homogeneous GNN models.

\begin{table}[htb]
    \caption{Model Performance of GAT by Attention Headers}
    \label{tb:gat-heads}
    \centering
    \begin{tabular}{l|r|r|r|r}
    \hline
        Model & Precision & Recall & F1 & PR-AUC \\ \hline
        GCN & \textbf{0.8817} & 0.4387 & 0.5843 & 0.7154 \\ \hline
        GAT (2 Heads) & 0.8695 & 0.5093 & 0.6399 & 0.7558 \\ 
        GAT (3 Heads) & 0.8815 & 0.4896 & 0.6304 & 0.7542 \\ 
        GAT (4 Heads) & 0.8559 & 0.5521 & 0.6670 & 0.7625 \\ \hline
        SAGE & 0.8244 & \textbf{0.6644} & \textbf{0.7354} & \textbf{0.8150} \\ \hline
    \end{tabular}
\end{table}


Table \ref{tb:units-sage} shows the model performance of the GraphSAGE model by the hidden layer size (the number of hidden units). With a larger hidden layer size, all model performance metrics gradually improved. In particular, the recall values with the largest hidden layer (256 units) are more than 10\% larger than the smallest layer (16 units). The F1-score and PR-AUC are also improved by 7\% and 4.5\%, respectively.

\begin{table}[htb]
    \caption{GraphSAGE Model Performance (Hidden Layers)}
    \label{tb:units-sage}
    \centering
    \begin{tabular}{l|r|r|r|r}
    \hline
        \# Units & Precision & Recall & F1 & PR-AUC \\ \hline
        16 & 0.8008 & 0.5833 & 0.6805 & 0.7797 \\ 
        32 & 0.8061 & 0.6181 & 0.6990 & 0.7874 \\ 
        64 & 0.8160 & 0.6377 & 0.7155 & 0.8051 \\ 
        128 & 0.8318 & 0.6620 & 0.7344 & 0.8174 \\ 
        256 & \textbf{0.8378} & \textbf{0.6840} & \textbf{0.7513} & \textbf{0.8250} \\ \hline
    \end{tabular}
\end{table}

Table \ref{tb:rl-sage} shows the model performance of the GraphSAGE model by the learning rate. With more effective learning rates, all metrics consistently decreased. Remarkably, the recall and F1-score drastically dropped in large learning rate more than 0.01.

\begin{table}[htb]
    \caption{GraphSAGE Model Performance (Learning Rate)}
    \label{tb:rl-sage}
    \centering
    \begin{tabular}{l|r|r|r|r}
    \hline
        Learning Rate & Precision & Recall & F1 & PR-AUC \\ \hline
        0.001 & 0.8303 & \textbf{0.6690} & 0.7381 & 0.8069 \\ 
        0.002 & 0.8256 & 0.6667 & \textbf{0.7389} & 0.8070 \\ 
        0.005 & \textbf{0.8336} & 0.6597 & 0.7387 & \textbf{0.8177} \\ 
        0.01 & 0.8257 & 0.6632 & 0.7350 & 0.8130 \\ 
        0.02 & 0.8268 & 0.6470 & 0.7218 & 0.8080 \\ 
        0.05 & 0.8195 & 0.6285 & 0.7119 & 0.8003 \\ 
        0.1 & 0.8002 & 0.5938 & 0.6825 & 0.7811 \\ \hline
    \end{tabular}
\end{table}

Table \ref{tb:dropout-sage} shows the performance of the GraphSAGE model by the dropout ratio. The trend of model performance with a dropout ratio is quite similar to the learning rate; the recall and F1-score values have drastically dropped with a larger dropout ratio.

\begin{table}[htb]
    \caption{GraphSAGE Model Performance (Dropout Ratio)}
    \label{tb:dropout-sage}
    \centering
    \begin{tabular}{l|r|r|r|r}
    \hline
        Dropout Ratio & Precision & Recall & F1 & PR-AUC \\ \hline
        No Dropout & \textbf{0.8377} & \textbf{0.7049} & \textbf{0.7659} & \textbf{0.8401} \\ 
        0.25 & 0.8335 & 0.6852 & 0.7514 & 0.8273 \\ 
        0.50 & 0.8298 & 0.6655 & 0.7374 & 0.8151 \\ 
        0.75 & 0.8155 & 0.6215 & 0.7054 & 0.7943 \\ \hline
    \end{tabular}
\end{table}

Table \ref{tb:wd-sage} shows the model performance of the GraphSAGE model in weight decay. It was a more noticeable trend than the learning rate and the dropout ratio. The precision and recall values were 87.5\% and 74.4\%, respectively, without weight decay. With 0.005 weight decay, however, the precision and recall values have dropped by 10\% and 19\%, respectively. Dropout and weight decay techniques usually prevent overfitting, but these were not effective in this application.


\begin{table}[htb]
    \caption{GraphSAGE Model Performance (Weight Decay)}
    \label{tb:wd-sage}
    \centering
    \begin{tabular}{l|r|r|r|r}
    \hline
        Weight Decay & Precision & Recall & F1 & PR-AUC \\ \hline
        No Weight Decay & \textbf{0.8750} & \textbf{0.7442} & \textbf{0.8021} & \textbf{0.8641} \\ 
        0.0001 & 0.8650 & 0.6944 & 0.7678 & 0.8404 \\ 
        0.0005 & 0.8144 & 0.6690 & 0.7347 & 0.8155 \\ 
        0.001 & 0.8000 & 0.6343 & 0.7045 & 0.7931 \\ 
        0.005 & 0.7758 & 0.5556 & 0.6478 & 0.7576 \\ \hline
    \end{tabular}
\end{table}

Based on these results, we recommend that hyperparameters be set according to the following policy for phishing fraud detection using the GNN model.
\begin{enumerate}
    \item GAT heads have little impact on model performance, and better results are obtained with GraphSAGE.
    \item The size of the hidden layer should be large.
    \item The learning rate should be small.
    \item Dropout and weight decay usually reduce overfitting, should be avoided as they adversely affect the overall performance.
\end{enumerate}

\section{Discussion}

\subsection{Homogeneous vs Heterogeneous GNN Models}

The results show that GNN models considering the node pair or edge importance yielded higher model performance than the baseline GCN. When comparing homogeneous GNN models, respectively, GCN, GAT, and GraphSAGE had higher values for recall, F1-score, and PR-AUC, in that order. Concerning precision, the GAT model is the highest among these, followed by GCN and GraphSAGE, but Recall, F1-score, and PR-AUC are more important for algorithms that detect very few phishing accounts.

RGCN, in which the GCN models were applied separately according to edge type (combination of node types at both ends), outperformed other homogeneous GNN models in all measures of precision, recall, F1-score, and PR-AUC. The other heterogeneous GNN models, HAN and HGT models, also achieved higher recall, F1-score, and PR-AUC than the baseline GCN model and other homogeneous GNN models. We conclude that using the RGCN model with heterogeneous edge types is more practical based on these results.

\subsection{Heterogeneous Graph and GNN Models}

Even when comparing the heterogeneous GNN models RGCN, HAN, and HGT (without relative temporal encoding), RGCN consistently achieved the best model performance. HAN learns graph attention and GAT and has a similar trend with relatively high precision values. On the other hand, HGT had a relatively high recall and F1-score, similar to GraphSAGE. We hypothesized that the HAN and HGT models, which define a heterogeneous meta-path and learn attention, would make more accurate predictions. However, the RGCN model, which applied the GCN model by edge type, was the best on both measures.

The hypothesis is that by converting each node type into a feature vector of one-hot encoding, it is possible to obtain the same expressive power as RGCN. The one-hot encoding can be applied to GCN and GraphSAGE models with baseline features that do not have the one-hot encoding feature, and the results are similar to those of the RGCN. There are few improvements in model performance for any metrics and it does not reach the level of RGCN. The main possible reason is that the distribution of node types was mostly biased toward account types, and not enough information was obtained for node features.

\section{Conclusion}

In this study, we conducted a comprehensive set of comparative model performance evaluation experiments using representative homogeneous and heterogeneous GNN models to detect phishing fraud accounts against a real Ethereum trading network. From hypotheses based on the characteristics of the transaction log data, we constructed a heterogeneous graph concerning account (node) types. We also found that heterogeneous GNN models that consider transaction edge types perform better than homogeneous GNN models. In particular, the RGCN model, in which graph convolution was applied separately to each edge type, achieved the best model performance on all metrics. However, when node types were treated as new input features, the accuracy did not improve compared to the homogeneous GNN model as the baseline. In conclusion, the most effective way to detect phishing fraud with GNN models is to let the heterogeneous GNN model take advantage of the information in the transaction network.


Although we conducted comparative evaluation experiments based on heterogeneity considering node types, additional information may be helpful in phishing fraud detection in an existing transaction network, such as the time stamp at which the transaction occurred and the attributes present in each account. As future work, we will conduct further evaluation and optimization of model performance using a GNN model to incorporate this information. First, considering transaction timestamps may find patterns specific to fraudulent transactions more accurately. A GNN model that can handle dynamic changes in a graph structure, such as EvolveGCN, would be helpful. Second, we also suppose it is possible to predict phishing fraud transaction patterns by applying classification to a subgraph consisting of each account and accounts close to it. GNN models for graph classification tasks, such as GIN, can be suitable to solve this problem. We do not have sufficient datasets to construct such a dynamic graph and subgraph for graph classification. However, we expect that more sophisticated phishing detection models will be proposed in the future by combining these advanced GNN models.

\begin{acks}
This work was supported by JSPS KAKENHI Grant Numbers JP21K17749 and JP21K21280.
\end{acks}

\bibliographystyle{ACM-Reference-Format}
\bibliography{references}


\begin{thebibliography}{15}


\ifx \showCODEN    \undefined \def \showCODEN     #1{\unskip}     \fi
\ifx \showDOI      \undefined \def \showDOI       #1{#1}\fi
\ifx \showISBNx    \undefined \def \showISBNx     #1{\unskip}     \fi
\ifx \showISBNxiii \undefined \def \showISBNxiii  #1{\unskip}     \fi
\ifx \showISSN     \undefined \def \showISSN      #1{\unskip}     \fi
\ifx \showLCCN     \undefined \def \showLCCN      #1{\unskip}     \fi
\ifx \shownote     \undefined \def \shownote      #1{#1}          \fi
\ifx \showarticletitle \undefined \def \showarticletitle #1{#1}   \fi
\ifx \showURL      \undefined \def \showURL       {\relax}        \fi
\providecommand\bibfield[2]{#2}
\providecommand\bibinfo[2]{#2}
\providecommand\natexlab[1]{#1}
\providecommand\showeprint[2][]{arXiv:#2}

\bibitem[Chen et~al\mbox{.}(2020)]%
        {acmtran}
\bibfield{author}{\bibinfo{person}{Liang Chen}, \bibinfo{person}{Jiaying Peng},
  \bibinfo{person}{Yang Liu}, \bibinfo{person}{Jintang Li},
  \bibinfo{person}{Fenfang Xie}, {and} \bibinfo{person}{Zibin Zheng}.}
  \bibinfo{year}{2020}\natexlab{}.
\newblock \showarticletitle{Phishing Scams Detection in Ethereum Transaction
  Network}.
\newblock \bibinfo{journal}{\emph{ACM Trans. Internet Technol.}}
  \bibinfo{volume}{21}, \bibinfo{number}{1}, Article \bibinfo{articleno}{10}
  (\bibinfo{date}{dec} \bibinfo{year}{2020}), \bibinfo{numpages}{16}~pages.
\newblock
\showISSN{1533-5399}
\urldef\tempurl%
\url{https://doi.org/10.1145/3398071}
\showDOI{\tempurl}


\bibitem[Hamilton et~al\mbox{.}(2017)]%
        {graphsage}
\bibfield{author}{\bibinfo{person}{Will Hamilton}, \bibinfo{person}{Zhitao
  Ying}, {and} \bibinfo{person}{Jure Leskovec}.}
  \bibinfo{year}{2017}\natexlab{}.
\newblock \showarticletitle{Inductive representation learning on large graphs}.
\newblock \bibinfo{journal}{\emph{Advances in neural information processing
  systems}}  \bibinfo{volume}{30} (\bibinfo{year}{2017}).
\newblock


\bibitem[Hu et~al\mbox{.}(2020)]%
        {hgt}
\bibfield{author}{\bibinfo{person}{Ziniu Hu}, \bibinfo{person}{Yuxiao Dong},
  \bibinfo{person}{Kuansan Wang}, {and} \bibinfo{person}{Yizhou Sun}.}
  \bibinfo{year}{2020}\natexlab{}.
\newblock \bibinfo{booktitle}{\emph{Heterogeneous Graph Transformer}}.
\newblock \bibinfo{publisher}{Association for Computing Machinery},
  \bibinfo{address}{New York, NY, USA}, \bibinfo{pages}{2704–2710}.
\newblock
\showISBNx{9781450370233}
\urldef\tempurl%
\url{https://doi.org/10.1145/3366423.3380027}
\showURL{%
\tempurl}


\bibitem[Kipf and Welling(2016)]%
        {gcn}
\bibfield{author}{\bibinfo{person}{Thomas~N Kipf} {and} \bibinfo{person}{Max
  Welling}.} \bibinfo{year}{2016}\natexlab{}.
\newblock \showarticletitle{Semi-supervised classification with graph
  convolutional networks}.
\newblock \bibinfo{journal}{\emph{arXiv preprint arXiv:1609.02907}}
  (\bibinfo{year}{2016}).
\newblock


\bibitem[Lin et~al\mbox{.}(2020)]%
        {t-edge}
\bibfield{author}{\bibinfo{person}{Dan Lin}, \bibinfo{person}{Jiajing Wu},
  \bibinfo{person}{Qi Yuan}, {and} \bibinfo{person}{Zibin Zheng}.}
  \bibinfo{year}{2020}\natexlab{}.
\newblock \showarticletitle{T-edge: Temporal weighted multidigraph embedding
  for ethereum transaction network analysis}.
\newblock \bibinfo{journal}{\emph{Frontiers in Physics}}  \bibinfo{volume}{8}
  (\bibinfo{year}{2020}), \bibinfo{pages}{204}.
\newblock


\bibitem[{NVIDIA}(2022)]%
        {dgx1}
\bibfield{author}{\bibinfo{person}{{NVIDIA}}.} \bibinfo{year}{2022}\natexlab{}.
\newblock \bibinfo{booktitle}{\emph{{Essential Instrument for AI Research |
  NVIDIA DGX-1}}}.
\newblock
\urldef\tempurl%
\url{https://www.nvidia.com/en-us/data-center/dgx-1/}
\showURL{%
\tempurl}


\bibitem[Schlichtkrull et~al\mbox{.}(2018)]%
        {rgcn}
\bibfield{author}{\bibinfo{person}{Michael Schlichtkrull},
  \bibinfo{person}{Thomas~N. Kipf}, \bibinfo{person}{Peter Bloem},
  \bibinfo{person}{Rianne van den Berg}, \bibinfo{person}{Ivan Titov}, {and}
  \bibinfo{person}{Max Welling}.} \bibinfo{year}{2018}\natexlab{}.
\newblock \showarticletitle{Modeling Relational Data with Graph Convolutional
  Networks}. In \bibinfo{booktitle}{\emph{The Semantic Web}},
  \bibfield{editor}{\bibinfo{person}{Aldo Gangemi}, \bibinfo{person}{Roberto
  Navigli}, \bibinfo{person}{Maria-Esther Vidal}, \bibinfo{person}{Pascal
  Hitzler}, \bibinfo{person}{Rapha{\"e}l Troncy}, \bibinfo{person}{Laura
  Hollink}, \bibinfo{person}{Anna Tordai}, {and} \bibinfo{person}{Mehwish
  Alam}} (Eds.). \bibinfo{publisher}{Springer International Publishing},
  \bibinfo{address}{Cham}, \bibinfo{pages}{593--607}.
\newblock
\showISBNx{978-3-319-93417-4}


\bibitem[Srivastava et~al\mbox{.}(2014)]%
        {dropout}
\bibfield{author}{\bibinfo{person}{Nitish Srivastava},
  \bibinfo{person}{Geoffrey Hinton}, \bibinfo{person}{Alex Krizhevsky},
  \bibinfo{person}{Ilya Sutskever}, {and} \bibinfo{person}{Ruslan
  Salakhutdinov}.} \bibinfo{year}{2014}\natexlab{}.
\newblock \showarticletitle{Dropout: a simple way to prevent neural networks
  from overfitting}.
\newblock \bibinfo{journal}{\emph{The journal of machine learning research}}
  \bibinfo{volume}{15}, \bibinfo{number}{1} (\bibinfo{year}{2014}),
  \bibinfo{pages}{1929--1958}.
\newblock


\bibitem[Vaswani et~al\mbox{.}(2017)]%
        {transformer}
\bibfield{author}{\bibinfo{person}{Ashish Vaswani}, \bibinfo{person}{Noam
  Shazeer}, \bibinfo{person}{Niki Parmar}, \bibinfo{person}{Jakob Uszkoreit},
  \bibinfo{person}{Llion Jones}, \bibinfo{person}{Aidan~N. Gomez},
  \bibinfo{person}{\L{}ukasz Kaiser}, {and} \bibinfo{person}{Illia
  Polosukhin}.} \bibinfo{year}{2017}\natexlab{}.
\newblock \showarticletitle{Attention is All You Need}. In
  \bibinfo{booktitle}{\emph{Proceedings of the 31st International Conference on
  Neural Information Processing Systems}} (Long Beach, California, USA)
  \emph{(\bibinfo{series}{NIPS'17})}. \bibinfo{publisher}{Curran Associates
  Inc.}, \bibinfo{address}{Red Hook, NY, USA}, \bibinfo{pages}{6000–6010}.
\newblock
\showISBNx{9781510860964}


\bibitem[Velickovic et~al\mbox{.}(2017)]%
        {gat}
\bibfield{author}{\bibinfo{person}{Petar Velickovic}, \bibinfo{person}{Guillem
  Cucurull}, \bibinfo{person}{Arantxa Casanova}, \bibinfo{person}{Adriana
  Romero}, \bibinfo{person}{Pietro Lio}, {and} \bibinfo{person}{Yoshua
  Bengio}.} \bibinfo{year}{2017}\natexlab{}.
\newblock \showarticletitle{Graph attention networks}.
\newblock \bibinfo{journal}{\emph{stat}}  \bibinfo{volume}{1050}
  (\bibinfo{year}{2017}), \bibinfo{pages}{20}.
\newblock


\bibitem[Wang et~al\mbox{.}(2021)]%
        {tsgn}
\bibfield{author}{\bibinfo{person}{Jinhuan Wang}, \bibinfo{person}{Pengtao
  Chen}, \bibinfo{person}{Shanqing Yu}, {and} \bibinfo{person}{Qi Xuan}.}
  \bibinfo{year}{2021}\natexlab{}.
\newblock \showarticletitle{TSGN: Transaction Subgraph Networks for Identifying
  Ethereum Phishing Accounts}. In \bibinfo{booktitle}{\emph{Blockchain and
  Trustworthy Systems}}, \bibfield{editor}{\bibinfo{person}{Hong-Ning Dai},
  \bibinfo{person}{Xuanzhe Liu}, \bibinfo{person}{Daniel~Xiapu Luo},
  \bibinfo{person}{Jiang Xiao}, {and} \bibinfo{person}{Xiangping Chen}} (Eds.).
  \bibinfo{publisher}{Springer Singapore}, \bibinfo{address}{Singapore},
  \bibinfo{pages}{187--200}.
\newblock
\showISBNx{978-981-16-7993-3}


\bibitem[Wang et~al\mbox{.}(2019b)]%
        {dgl}
\bibfield{author}{\bibinfo{person}{Minjie Wang}, \bibinfo{person}{Da Zheng},
  \bibinfo{person}{Zihao Ye}, \bibinfo{person}{Quan Gan},
  \bibinfo{person}{Mufei Li}, \bibinfo{person}{Xiang Song},
  \bibinfo{person}{Jinjing Zhou}, \bibinfo{person}{Chao Ma},
  \bibinfo{person}{Lingfan Yu}, \bibinfo{person}{Yu Gai},
  \bibinfo{person}{Tianjun Xiao}, \bibinfo{person}{Tong He},
  \bibinfo{person}{George Karypis}, \bibinfo{person}{Jinyang Li}, {and}
  \bibinfo{person}{Zheng Zhang}.} \bibinfo{year}{2019}\natexlab{b}.
\newblock \showarticletitle{Deep Graph Library: A Graph-Centric,
  Highly-Performant Package for Graph Neural Networks}.
\newblock \bibinfo{journal}{\emph{arXiv preprint arXiv:1909.01315}}
  (\bibinfo{year}{2019}).
\newblock


\bibitem[Wang et~al\mbox{.}(2019a)]%
        {han}
\bibfield{author}{\bibinfo{person}{Xiao Wang}, \bibinfo{person}{Houye Ji},
  \bibinfo{person}{Chuan Shi}, \bibinfo{person}{Bai Wang},
  \bibinfo{person}{Yanfang Ye}, \bibinfo{person}{Peng Cui}, {and}
  \bibinfo{person}{Philip~S Yu}.} \bibinfo{year}{2019}\natexlab{a}.
\newblock \showarticletitle{Heterogeneous Graph Attention Network}. In
  \bibinfo{booktitle}{\emph{The World Wide Web Conference}} (San Francisco, CA,
  USA) \emph{(\bibinfo{series}{WWW '19})}. \bibinfo{publisher}{Association for
  Computing Machinery}, \bibinfo{address}{New York, NY, USA},
  \bibinfo{pages}{2022–2032}.
\newblock
\showISBNx{9781450366748}
\urldef\tempurl%
\url{https://doi.org/10.1145/3308558.3313562}
\showDOI{\tempurl}


\bibitem[Wu et~al\mbox{.}(2020)]%
        {trans2vec}
\bibfield{author}{\bibinfo{person}{Jiajing Wu}, \bibinfo{person}{Qi Yuan},
  \bibinfo{person}{Dan Lin}, \bibinfo{person}{Wei You}, \bibinfo{person}{Weili
  Chen}, \bibinfo{person}{Chuan Chen}, {and} \bibinfo{person}{Zibin Zheng}.}
  \bibinfo{year}{2020}\natexlab{}.
\newblock \showarticletitle{Who are the phishers? phishing scam detection on
  ethereum via network embedding}.
\newblock \bibinfo{journal}{\emph{IEEE Transactions on Systems, Man, and
  Cybernetics: Systems}} (\bibinfo{year}{2020}).
\newblock


\bibitem[Xie et~al\mbox{.}(2021)]%
        {tasmg}
\bibfield{author}{\bibinfo{person}{Yunyi Xie}, \bibinfo{person}{Jie Jin},
  \bibinfo{person}{Jian Zhang}, \bibinfo{person}{Shanqing Yu}, {and}
  \bibinfo{person}{Qi Xuan}.} \bibinfo{year}{2021}\natexlab{}.
\newblock \showarticletitle{Temporal-Amount Snapshot MultiGraph for Ethereum
  Transaction Tracking}. In \bibinfo{booktitle}{\emph{Blockchain and
  Trustworthy Systems}}, \bibfield{editor}{\bibinfo{person}{Hong-Ning Dai},
  \bibinfo{person}{Xuanzhe Liu}, \bibinfo{person}{Daniel~Xiapu Luo},
  \bibinfo{person}{Jiang Xiao}, {and} \bibinfo{person}{Xiangping Chen}} (Eds.).
  \bibinfo{publisher}{Springer Singapore}, \bibinfo{address}{Singapore},
  \bibinfo{pages}{133--146}.
\newblock
\showISBNx{978-981-16-7993-3}


\end{thebibliography}


\end{document}